\documentclass[sn-mathphys-num]{sn-jnl}

\usepackage{graphicx}%
\usepackage{multirow}%
\usepackage{amsmath,amssymb,amsfonts}%
\usepackage{amsthm}%
\usepackage{mathrsfs}%
\usepackage[title]{appendix}%
\usepackage{xcolor}%
\usepackage{textcomp}%
\usepackage{manyfoot}%
\usepackage{booktabs}%
\usepackage{algorithm}%
\usepackage{algorithmicx}%
\usepackage{algpseudocode}%
\usepackage{listings}%
\usepackage{array}

\begin{document}
\pagenumbering{arabic}
\setcounter{page}{1}
\setcounter{figure}{0}
\setcounter{table}{0}
\renewcommand{\thefigure}{\arabic{figure}}
\renewcommand{\thefootnote}{\arabic{footnote}}

\title{Reflecting Topology Consistency and Abnormality via Learnable Attentions for Airway Labeling}


\author[1,2]{\fnm{Chenyu} \sur{Li}}\email{lichenyu@sjtu.com}

\author[1,2]{\fnm{Minghui} \sur{Zhang}}\email{minghuizhang@sjtu.edu.cn}
\author[1,2]{\fnm{Chuyan} \sur{Zhang}}\email{zhangchuyan@sjtu.edu.cn}
\author*[1,2]{\fnm{Yun} \sur{Gu}}\email{yungu@ieee.org}

\affil[1]{\orgdiv{Institute of Medical Robotics}, \orgname{Shanghai Jiao Tong University}, \orgaddress{\city{Shanghai}, \country{China}}}

\affil[2]{\orgdiv{Institute of Image Processing and Pattern Recognition}, \orgname{Shanghai Jiao Tong University}, \orgaddress{\city{Shanghai}, \country{China}}}


\abstract{\textbf{Purpose:} 
Accurate airway anatomical labeling is crucial for clinicians to identify and navigate complex bronchial structures during bronchoscopy. 
Automatic airway labeling is challenging due to significant anatomical variations. 
Previous methods are prone to generate inconsistent predictions,hindering preoperative planning and intraoperative navigation.
This paper aims to enhance topological consistency and improve the detection of abnormal airway branches.

\textbf{Methods:}
We propose a transformer-based framework incorporating two modules: the Soft Subtree Consistency (SSC) and the Abnormal Branch Saliency (ABS). 
The SSC module constructs a soft subtree to capture clinically relevant topological relationships, allowing for flexible feature aggregation within and across subtrees. 
The ABS module facilitates interaction between node features and prototypes to distinguish abnormal branches, preventing the erroneous features aggregation between normal and abnormal nodes.

\textbf{Results:} Evaluated on a challenging dataset characterized by severe airway deformities, our method achieves superior performance compared to state-of-the-art approaches. 
Specifically, it attains an 83.7\% subsegmental accuracy, along with a 3.1\% increase in segmental subtree consistency, a 45.2\% increase in abnormal branch recall. Notably, the method demonstrates robust performance in cases with airway deformities, ensuring consistent and accurate labeling.

\textbf{Conclusion:}  The enhanced topological consistency and robust identification of abnormal branches provided by our method offer an accurate and robust solution for airway labeling, 
with potential to improve the precision and safety of bronchoscopy procedures. \footnote{The code is publicly available at \href{https://github.com/EndoluminalSurgicalVision-IMR/Reflecting-Topology-Consistency-and-Abnormality-via-Learnable-Attentions}{Reflecting-Topology-Consistency-and-Abnormality-via-Learnable-Attentions} } }

\keywords{Airway anatomical labeling, Structural prior, Anomaly detection, Transformer}



\maketitle

\section{Introduction}\label{sec1}
\begin{figure}[!t]
\centering
\includegraphics[width=0.9\textwidth]{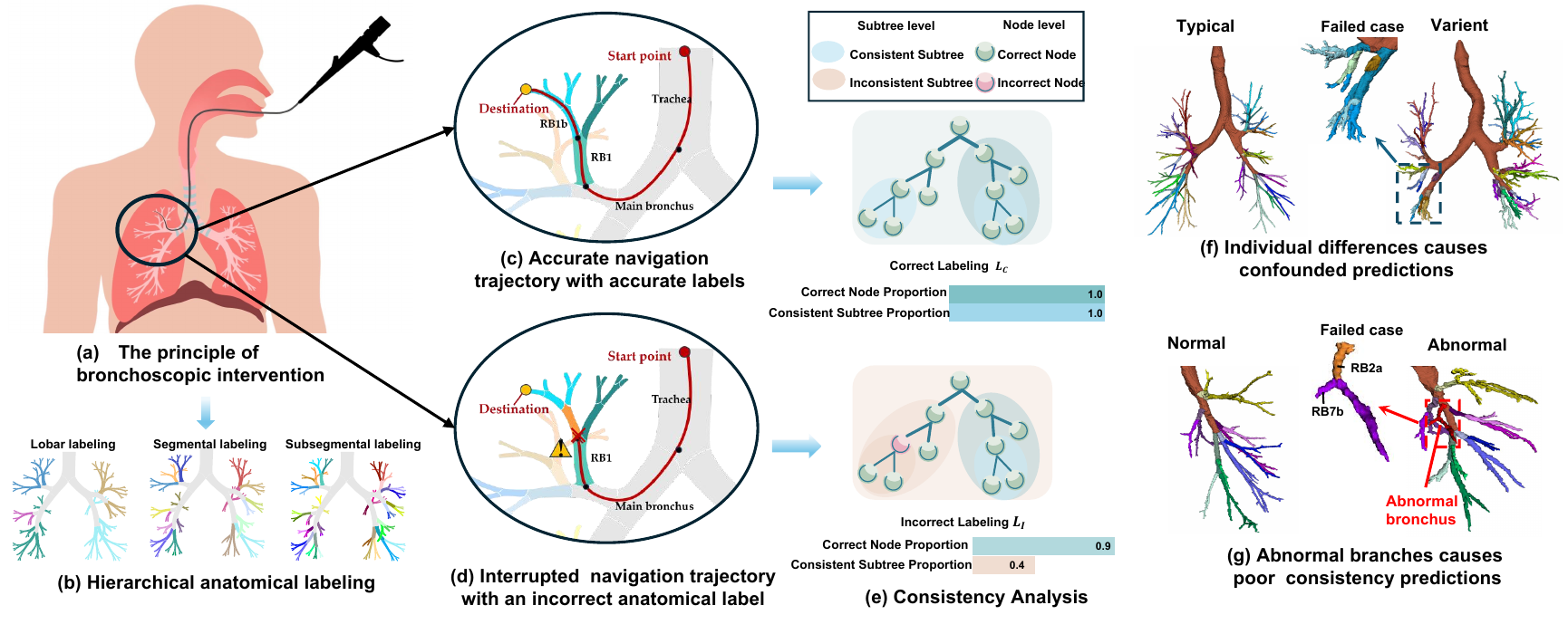}
\caption{Overview of bronchial anatomical labeling and its challenges.
(a) and (b) illustrate the preoperative planning path for bronchoscopy guided by hierarchical anatomical labeling.
(c) and (d) show successful and failed navigation trajectories, respectively.
(e) compares node-wise correctness and subtree-wise consistency under correct labeling ($L_C$) and incorrect labeling ($L_I$). The results indicate that anatomical subtree-wise consistency is more sensitive to labeling errors.
(f) and (g) demonstrate how individual variability and abnormal branches pose challenges to labeling accuracy and consistency} \label{fig1}
\end{figure}

Bronchial anatomical labeling enables clinicians to precisely identify the bronchus or subsegment involved with the pathological lesion, 
thereby facilitating the optimal planning of the surgical approach while minimizing injury to surrounding healthy tissues~\cite{bib2}. 
When integrated with advanced navigation technologies, the hierarchical bronchial labeling (lobar, segmental, and subsegmental levels) supports the precise guidance of surgical instruments by robotic or navigational systems(as illustrated in  Fig.~\ref{fig1}(a) and (b)). This ensures accurate execution of procedures within the intricate pulmonary anatomy~\cite{bib0}.
Accurate anatomical labeling ensures reliable navigation trajectories for bronchoscopy; while labeling errors can lead to failed interventions, as shown in Fig.~\ref{fig1}(c) and (d).

In addition, bronchial anatomical labeling allows for a comprehensive assessment of resectability and improved identification of complicated anatomical variations for segmentectomy~\cite{bib25}. 
Furthermore, its applications extend beyond procedural applications to include diagnostic advancements~\cite{bib14}. These features serve as robust biomarkers for clinical diagnosis and prognosis.

These clinical applications focus on local segments and subsegments surrounding the lesions, underscoring the importance of ensuring the precision and completeness of each anatomical region. As shown in Fig.~\ref{fig1}, these regions can naturally be represented as anatomical subtrees within the airway structure. Unlike branch-wise correctness, anatomical subtree-wise consistency reflects the integrity of the entire anatomical structure, as demonstrated in  Fig.~\ref{fig1}(e), which is critical to deliver precise and clinically meaningful outcomes.



Manual annotation of bronchial anatomical labeling by medical experts is time-consuming and requires a high level of expertise. Different from the binary segmentation tasks~\cite{bib21}, this task assigns the branches with specific class labels according to the anatomy. Therefore, automated labeling algorithms have emerged to alleviate the workload of clinicians. However, automatic bronchial anatomical labeling among different patients can be challenging from two aspects. 
Firstly, individual variability leads to substantial feature overlap, further aggravated by disease-induced airway deformities.
Traditional algorithms rely on predefined rules~\cite{bib2,bib3} or association graph search~\cite{bib4,bib5}, 
failing to accurately label distal branches. Recently, deep learning methods~\cite{bib9,bib10,bib16} have demonstrated superior performance on this task. 
To address individual differences, Yu \textit{et al.} \cite{bib9} introduced a hypergraph model to encode subtree representations through hyperedges. However, it relies heavily on manually designed hypergraphs and is sensitive to hyperparameters. Xie \textit{et al.} \cite{bib16} encoded deep point-graph representation for analyzing 3D pulmonary tree structures, but only achieved segmental level classification. AirwayFormer~\cite{bib10} was proposed to aggregate features using a self-attention mechanism and neighborhood encoding. Although these approaches achieved promising labeling results, limited attention has been paid to airway deformity such as atrophy, compression, and dilatation, resulting in failure in deformed airways as shown in Fig.~\ref{fig1}(f).
Secondly, the bronchial tree exhibits considerable morphological variability, with many anatomical variations beyond established nomenclatures. 
These abnormal branches without certain locations and orientations lack consistent anatomical descriptions, which complicates identification and impacts 
the classification of other branches. However, current methods often overlook these variant branches. Huang \textit{et al.} \cite{bib15} treated abnormal bronchus as a separate category, while the absence of anatomical standards made it challenging to categorize them equally with other classes as illustrated in Fig.~\ref{fig1}(g).

To alleviate the aforementioned challenges, we focus on the topological consistency for bronchial anatomical labeling. 
Specifically, we propose the Soft Subtree Consistency (SSC) module to tackle the first challenge,
SSC constructs a soft subtree via feature interaction and topological refinement. Segmental labels serve as the criterion for soft subtrees, 
resulting in more flexible and discriminative feature aggregation within and across subtrees. 
To address the second challenge,  we propose an Abnormal Branch Saliency (ABS) module, which compares node features and prototypes to predict anomaly scores, distinguishing abnormal nodes. And a soft mask derived from anomaly scores is utilized to prevent feature aggregation between normal and abnormal nodes for a smoother regularization. 

Building on a transformer-based framework, we introduce anatomical subtree-wise consistency as an evaluation standard to better reflect the integrity of bronchial structures. Compared to existing methods that focus on branch-wise labeling, our approach achieves notable improvements under this clinically relevant metric, particularly in complex or variant airway morphologies.

The proposed method is evaluated on ATL-fibrosis dataset, a challenging dataset we constructed to facilitate comprehensive evaluation of airway labeling methods.
Extensive experiments demonstrate that our method achieves superior performance compared to other state-of-the-art methods, 
achieving 91.4\% accuracy in segmental, 83.7\% accuracy in subsegmental with a 1.4\% increase, 96.8\% consistency with a 3.1\% increase. The results demonstrate robustness across a diverse range of airway conditions, including severe abnormalities and anatomical variations, relevant to clinical scenarios.

\section{Method}\label{sec2}
\begin{figure}[!t]
\centering
\includegraphics[width=0.9\textwidth]{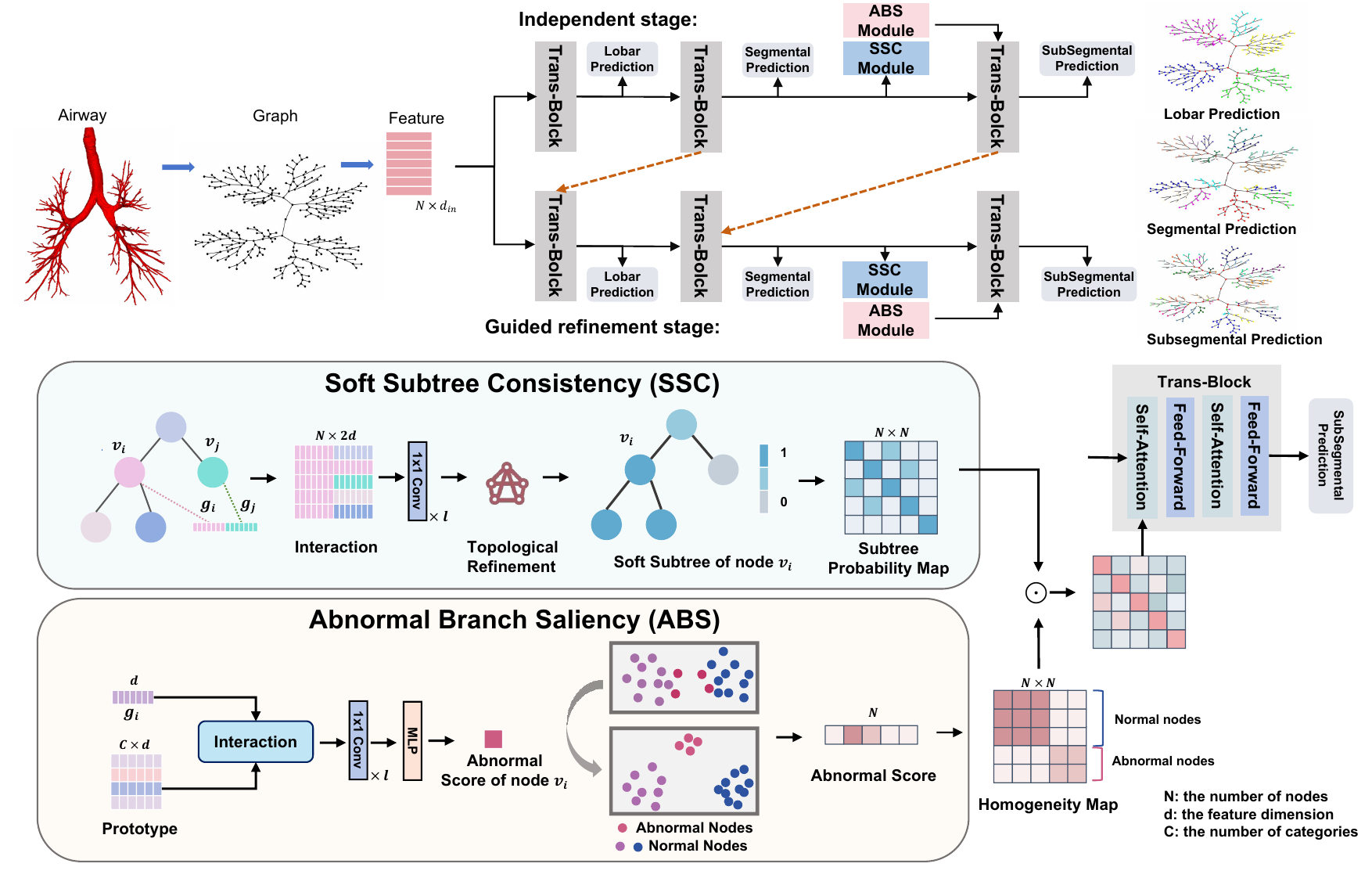}
\caption{Overview of the proposed method: The airway labeling method comprises a Soft Subtree Consistency (SSC) module, which generates a soft subtree probability map, and an Abnormal Branch Saliency (ABS) module, which produces an abnormality score prediction. Both of them are used to inform attention interaction at the subsegmental level}\label{fig2}
\end{figure}
Each airway branch is represented as a node in the graph, with attributes designed to reflect its location, morphology, and structure. An overview of the proposed method is illustrated in Fig.~\ref{fig2}. We propose two modules to improve the consistency of prediction: (1) SSC module (detailed in Section \ref{subsec22}) softly encodes the relationship between subtrees. (2) ABS module (detailed in Section \ref{subsec23}) distinguishes between normal nodes and abnormal nodes.
\subsection{Overall Framework}\label{subsec21}
We adopt a U-shaped framework that progressively improves classification accuracy by incorporating hierarchical relationships. This framework comprises two stages: an independent stage ($1st$ stage) and a guided refinement stage ($2nd$ stage). The latter employs a fine-to-coarse guidance mechanism, using finer features from the independent stage to refine coarser predictions, thereby enhancing overall labeling performance.

In the i-th ($i = 1, 2$) stage, where the 1st one denotes the independent stage, and the 2nd one denotes the guided refinement stage, airway classification tasks are refined through cascaded transformer blocks:
\begin{align}
    &\tilde{G}_m^i = \mathcal{F}(q=k=v=X_m^i),  \\
    &G_m^i = \left\{
    \begin{array}{lll}
        \mathcal{F}(q = \tilde{G}_m^i, k = v = G_{m+1}^{i-1}) & & \text{for } m \in \{lob, seg\} \text{ and } i = 2 \\
        \tilde{G}_m^i & & \text{otherwise}
    \end{array} \right.\\
    & X_{m+1}^i = G_m^i \quad \text{ for } m \in \{lob, seg\}   \\
    & P_m^i = \mathcal{C}_m^i(G_m). 
\end{align}
Here, the hierarchical level $m \in \mathcal{H}$, where the ordered set $\mathcal{H} = \{lob, seg, sub\} $ corresponding to lobar, segmental, and subsegmental levels respectively. $\mathcal{F}$ denotes the function of 
Transformer architecture. $X_m^i \in \mathbb{R}^{N\times d}$ and $G_m^i \in \mathbb{R}^{N\times d}$  represent the input and output features of the m-th transformer block in the i-th stage, where $N$ denotes the number of nodes and $d$ is the feature dimension. 
$\mathcal{C}_m^i$  denotes the m-th classifier consisting of a simple linear layer, while $P_m^i \in \mathbb{R}^{N\times c_m}$ is the prediction result where $c_{m}$ denotes the number of categories. 
For simplicity, we take the guided refinement stage for an example, omitting the index $i$ in subsequent notation.

The Transformer Block serves as the backbone and aggregates information globally through the attention mechanism~\cite{bib11}. A graph bias term~\cite{bib12} is added to the attention map, incorporating topological information:

\begin{equation} \label{eqA}
    A_m = \frac{(X_m Q_m)(X_m K_m)^T}{\sqrt{d}} + B_m ,
\end{equation} \\
$Q_m \in \mathbb{R}^{d \times d_k}$ and $K_m \in \mathbb{R}^{d \times d_k}$ denote the learnable attention parameters. We use $\psi(v_i, v_j)$ as the shortest path distance (SPD) between node $v_i$ and node $v_j$. The bias matrix $B_m \in \mathbb{R}^{N \times N}$ is a learnable codebook indexed by $\psi\in \mathbb{R}^{N \times N}$.

The finer classification utilizes the output of the coarser level as input, inheriting shared information to reduce the classification difficulty. 
Additionally, due to the corresponding relationship of the hierarchical nomenclature, finer predictions can directly infer coarser ones. 
Thus, finer features from the independent stage serve as additional input to guide coarse-grained classification. 

\subsection{ Soft Subtree Consistency}\label{subsec22}
Due to significant inter-individual variability, relying solely on node representations for classification can lead to substantial feature overlap. 
To address this challenge, we propose the SSC module, which dynamically encodes subtree representations, unlike hard-subtree techniques which rely on descendant counts for subtree construction.

We utilize segmental labels to define subtrees that convey clinically relevant anatomical meaning.
The soft subtree map $M_{t} \in \mathbb{R}^{N \times N}$ is introduced, where $M_{t}(i,j)$ represents the probability that nodes $v_i$ and $v_j$ belong to the same subtree. The ground truth labels for $M_{t}$ are defined as a binary map that represents the probability of two nodes sharing the same segmental label. Specially, $M(i,j)$ is 1 if node $v_j$ is a descendant of node $v_i$, and 0 otherwise. 

Soft subtree map $\Bar{M}_t$ is inferred through  segmental feature $G_{seg} =(g_1, g_2, \dots, g_N)^T \in \mathbb{R}^{N\times d}$, where $g_i$ represents the feature of node $v_i$. The inference involves each node feature $g_i$ interacting globally with all other nodes to capture intra-subtree relationships, followed by dimensionality reduction using learnable $1 \times 1$ convolutional layers(as shown in Fig.~\ref{fig2} (a)).
Then referencing the hard subtree which consists of a node and its descendants, the probability of two nodes belonging to the same subtree is increased if one node is a descendant of the other. To achieve this, a soft topological refinement is applied, where descendant relationships $M_D$ dynamically adjust the probabilities in $M_t$:
\begin{equation}
          \hat{M}_{t} = M_D \odot(\Bar{M}_{t} + K \odot (1- \Bar{M}_{t})) + (1-M_D) \odot \Bar{M}_{t},
\end{equation}

where $K \in [0,1]^{N \times N}$ is obtained from $G_{seg}$ to modulate the refinement. $M_D \in \mathbb{R}^{N\times N} $ is a binary mask that indicates the descendant relationship between nodes. Specifically, $M_D(i,j)$ is 1 if node $v_j$ is a descendant of node $v_i$, and 0 otherwise. 

\begin{figure}[!t]
\centering
\includegraphics[width=0.9\textwidth]{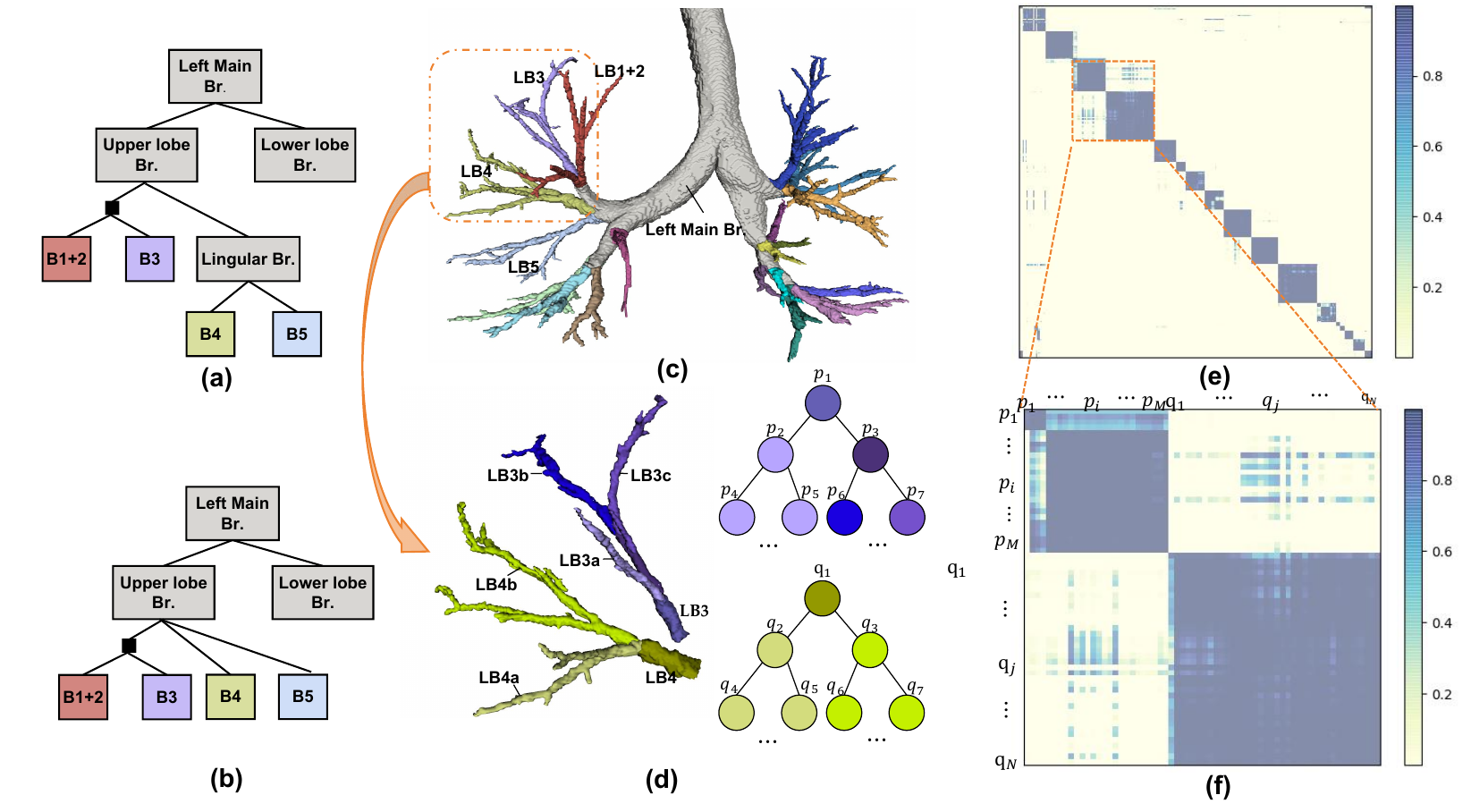}
\caption{ (a) illustrates the typical pattern of the left upper lobe, featuring a long lingular branch originating from the left main bronchus and bifurcating into LB4 and LB5. (b) shows a variant pattern where LB4 and LB5 directly bifurcates from the left main bronchus. (c) demonstrates a variant airway at the segmental level. (d) describes LB3 and LB4 at the subsegmental level. (e) shows the soft subtree map of the Airway. (f) details the Segment Consistency Mask} \label{fig3}
\end{figure}

According to the hierarchical nomenclature, the assignment of subsegmental labels is determined by their relative relationships to branches within a segment. Therefore, the soft subtree map is utilized to constrain feature aggregation within and across subtrees at the subsegmental level as follows,
\begin{equation}
     A_{sub} = \frac{(X_{sub}Q_{sub})(X_{sub}K_{sub})^T}{\sqrt{d}} \odot \hat{M}_{t},
\end{equation}

The soft subtree map subsequently promotes the clustering of feature representations according to shared subtree characteristics.
Subsequent attention layers calculated as Eq.~\eqref{eqA} with global receptive fields then refine these representations, allowing for distinctions between individual nodes. 
This approach improves the clustering of subsegmental features by segmental category and simplifies the prediction of subsegmental categories into encoding intra-subtree relations, effectively enhancing the accuracy and consistency of prediction.
 
 Feature overlap occurs when node representations from different anatomical pattern become indistinguishable, particularly in cases with substantial morphological variability. The SSC module proves particularly effective in handling anatomical variations.  For example, Fig.~\ref{fig3} (b-d) illustrates a scenario where the positions and orientations of LB4 branches in this variant differ significantly from the typical pattern (Fig.~\ref{fig3} (a)), resulting in substantial feature overlap between LB3 and LB4. The SSC module alleviates this issue by constraining feature aggregation within subtrees softly (as shown in Fig.~\ref{fig3} (e-f)), ensuring consistency even in cases of significant anatomical variability.  Moreover, by integrating topological refinement with attention mechanisms, it not only improves subtree clustering but also enhances overall anatomical coherence.

\subsection{Abnormal Branch Saliency}\label{subsec23}
 The position, orientation, and shape of abnormal branches are uncertain, with significant individual differences. The primary criterion for identifying an abnormal branch is its inability to be classified into any category within the nomenclature. Thus, we propose an ABS module that compares node features with class representations to detect anomalies.
First, the prototypes are obtained through the segmental features $G_{seg} \in \mathbb{R}^{N \times d}$ and prediction result $P_{seg} \in \mathbb{R}^{N \times c_{seg}}$ :
\begin{align}
      S_{seg} &= softmax(P_{seg}), \\
    H &= ((S_{seg})^\alpha)^T G_{seg},
\end{align}
$S_{seg} \in [0,1]^{N \times c_{seg}}$ denotes the probability matrix of the segmental level. $H \in \mathbb{R}^{c_{seg} \times d}$ are representations for segmental categories, while $\alpha$ is a learnable cluster parameter. Then they are refined through a transformer to incorporate global contextual information and ensure better class representations for each segmental category:
\begin{equation}
    \tilde{H} = \mathcal{F}(q = H,\ k=v=G_{seg}).
\end{equation}

After obtaining the refined class representations, node features interact with these prototypes $H$ to compute an anomaly score.
As shown in Fig.~\ref{fig2} (b), each pair of feature and prototypes is concatenated along the depth dimension, forming the  $\phi \in \mathbb{R}^{N \times c_{seg} \times 2d}$.  
$\phi$ is processed through $1\times 1$ convolutional layers to reduce the depth dimension, followed by the Multilayer Perceptron (MLP) to compute the anomaly score $\hat{Y}_{a} \in [0,1]^{N \times 1}$, indicating the probability of each node being an abnormal branch.

Due to the distinctive characteristics of abnormal branches, it is too absolute and arbitrary to use this anomaly score to directly classify nodes as a category equally 
with other classes. Instead, we generate a soft mask $M_{a} \in \mathbb{R}^{N \times N}$ for smoother regularization, 
which prevents feature aggregation between normal and abnormal nodes: 
\begin{equation}
    \hat{M}_{a} = 1 - (\hat{Y}_{a}\hat{Y}_{a}^T)^2.
\end{equation}
Mask $\hat{M}_{a}$ ensures that abnormal nodes are guided away from normal nodes in the feature space, reducing their disruptive influence on normal branch predictions.

To further enhance subsegmental feature interactions,  the combination between two soft masks $\hat{M}_{t} \odot \hat{M}_{a}$ , is applied to the attention interaction of subsegmental Transformer Block,
\begin{equation}\label{A2}
     A_{sub} = \frac{(X_{sub}Q_{sub})(X_{sub}K_{sub})^T}{\sqrt{d}} \odot (\hat{M}_{t} \odot \hat{M}_{a}),
\end{equation}

The ABS module not only identifies abnormal branches but also reduces their disruptive influence on normal branch consistency, indirectly enhancing the overall accuracy and robustness of the model. Additionally, the anomaly score provides valuable insights for clinicians to identify structural abnormalities and understand airway variations, improving the practical utility of the method.

For classification tasks, cross-entropy loss with label smoothing is used, excluding abnormal nodes.
Binary cross-entropy loss is employed for soft subtree map $\hat{M}_{t}$ prediction and abnormal 
score prediction $\hat{Y}_{a}$. The total loss function is formulated as:
\begin{equation}
\begin{split}
\mathcal{L} = \sum_{i=1}^{2} \gamma_{i} \Bigl( &\sum_{m\in \mathcal{H}} \alpha_m \mathcal{L}_{CE}(\hat{P}_m^i,Y_m) \\
&+ \beta_1 \mathcal{L}_{BCE}(\hat{M}_{t}^i,M_{t}) + \beta_2 \mathcal{L}_{BCE}(\hat{Y}_{a}^i,Y_{a}) \Bigr)
\end{split},
\end{equation}
where $\gamma_i$, $\alpha_m$, $\beta_1$, $\beta_2$ are empirically tuned parameters.

\section{EXPERIMENTS AND RESULTS}\label{sec3}

\subsection{Datasets and implementation details}\label{subsec31} 

We constructed ATL (Airway Tree Labeling) dataset by developing detailed subsegmental anatomical annotations encompassing up to 127 categories, based on chest CT-based binary segmentation mask from ATM22~\cite{bib21} and AIIB23~\cite{bib14}. Cases were selected based on visibility extending at least two generations beyond the subsegmental level to facilitate accurate airway labeling. The ATL-clean subset consisting of 196 cases derived from the ATM22 dataset, characterized by the absence of major airway deformities, was used for training and validation. For testing, the ATL-fibrosis subset includes 89 cases derived from AIIB23, a fibrosis airway dataset. This subset encompasses a diverse range of airways, including both normal anatomies and those with extensive deformities, enabling a comprehensive evaluation across both challenging and normal scenarios.

Following the pipeline proposed in TNN\cite{bib9}, airway centerlines were extracted from the binary segmentation results using the skeletonization algorithm \cite{bib17}, and bifurcation points were identified to divide the airway into branches. 
Branch-wise features were then automatically extracted based on the feature design outlined in TNN\cite{bib9}. 
A three level nomenclature~\cite{bib1} was adopted for the bronchial tree, comprising 6 types of lobar bronchi, 19 types of segmental bronchi, and 127 types of subsegmental bronchi, with an additional "outlier" category for abnormal nodes. Each segment could include three subsegments named ‘a’, ‘b’, and ‘c’ and the branch between bifurcations is named ‘a+b’, ‘b+c’ or ‘a+c’.

The weight of loss function $\gamma_i$, $\alpha_m$, $\beta_1$, and $\beta_2$ were set to 1. The label smoothing hyperparameter was set to 0.01. 
We employed the Adam optimizer for training with a learning rate of $5e^{-4}$ over 600 epochs. For lobar and segmental airway labeling, 
we stacked two transformers calculated as Eq.~\eqref{eqA}. At the subsegmental level, we stacked two transformers calculated as Eq.~\eqref{A2}, 
followed by two more as Eq.~\eqref{eqA}. 
Finally, the predictions from the guided refinement stage served as the output. Branches with an abnormal score greater than a threshold of 0.1 were predicted as abnormal branches. A sensitivity analysis of this threshold (see Appendix \hyperlink{secA1}{A}) confirms its robustness.
All experimental results are averaged over five runs with different random seeds to minimize the influence of randomness.  
\subsection{Evaluation metrics and results}\label{subsec32}

Four branch-wise evaluation metrics are used: accuracy (ACC), precision (PR), recall (RC), and F1 score (F1).
Additionally, two subtree-wise metrics were introduced: Subtree Consistency (SC), which evaluates the proportion of subtrees meeting anatomical labeling criteria, and Topological Distance (TD), which measures the shortest path length between predicted branches and their corresponding ground truth subtrees. The criteria for SC are:
\begin{enumerate}
    \item Segmental Consistency: All nodes in a subtree share the same predicted segmental label.
    \item  Subsegmental Consistency: All nodes in a subtree share the same predicted segmental label and conform to the defined subsegmental topology, with nodes labeled as ‘a’, ‘b’, ‘c’, or bifurcation branches (‘a+b’, ‘b+c’, or ‘a+c’).
\end{enumerate}
The evaluation metrics are formally defined as:
\begin{equation}
    SC = \frac{N_{cs}}{N_s} \qquad TD = \frac{1}{N}
\sum_{i=1}^{N} \min_{ j \in \{ {j|y_j = \hat{y}_i\}}} d_{(i,j)},
\end{equation}

where $N_s$ is the total number of subtrees and $N_{cs}$ denotes the consistent classified number. $N$ is the total number of nodes in a graph, and $d_{i,j}$ denotes the shortest path length between node $v_i$ and node $v_j$. $y_j$ and $\hat{y}_i$ denote the ground truth and prediction of node $v_i$ respectively.

\begin{table}[!h]
\caption{Quantitative comparison with different methods in airway anatomical labeling. PR, RC, and F1 are reported as percentages (\%). TD represents a unitless topological distance. 
}

\setlength{\tabcolsep}{1.6pt} 
\centering
\begin{tabular}{@{}c|c|c|cccc|ccc|cc|cc@{}}
\toprule
& \multicolumn{9}{c|}{Branch-wise} & \multicolumn{4}{c}{Subtree-wise}\\
\cmidrule(lr){2-10} \cmidrule(lr){11-14} 

 & \multicolumn{1}{c|}{Lob} & \multicolumn{1}{c|}{Seg} & \multicolumn{4}{c|}{Subsegmental} & \multicolumn{3}{c|}{Abnormal} & \multicolumn{2}{c|}{Segmental}& \multicolumn{2}{c}{Subsegmental}    \\
Method & ACC$\uparrow$ & ACC$\uparrow$ & ACC $\uparrow$& PR $\uparrow$& RC$\uparrow$ & F1 $\uparrow$& PR $\uparrow$& RC $\uparrow$& F1 $\uparrow$& SC$\uparrow$ & TD $\downarrow$& SC $\uparrow$& TD$\downarrow$ \\
\midrule
GCN\cite{bib18} &91.7& 67.3& 45.5& 30.5&  35.4& 29.5 & 0.0& 0.0& 0.0& 60.4 &2.17 & 43.9 & 4.04\\
GraphSAGE\cite{bib19} &93.3& 72.6&  55.7& 46.0&  49.8& 44.6 & 8.2 &2.7 & 3.6 & 64.4 & 1.86 & 45.3 & 3.35\\
UniSAGE\cite{bib23} &93.2& 72.9  &57.0& 46.9&  50.6& 45.6 &4.7 & 1.8 & 2.5 & 69.8 & 1.66 & 62.4 & 2.97\\
TNN\cite{bib9} &95.3 & 78.7 & 63.3& 50.6& 57.0&50.9 & 1.9 & 0.9 &1.1 & 85.4 & 1.31 & 73.2 & 2.50 \\
AirwayFormer\cite{bib10} &97.1 & 90.2& 82.3&78.1 & 79.6& 77.5 & 45.6 & 29.0 & 33.1 & 93.7 & 0.64 & 85.7 & 1.26 \\
DeepAssign\cite{bib20} &96.4 & 88.9& 81.0 & 75.3 & 77.5&74.8 &36.5 & 17.7 & 21.1 & 87.1 & 0.76 & 78.7 & 1.41  \\
Ours & 97.4  & \textbf{91.4}  & \textbf{83.7} & \textbf{79.1} & \textbf{80.9} & \textbf{78.6} & \textbf{55.6} & \textbf{62.9} & \textbf{54.8} & \textbf{96.8} & \textbf{0.54} & \textbf{86.3} & \textbf{1.17}\\
\botrule

\end{tabular}

\label{tab1}
\footnotetext{Bold values indicate the best result, and results for "Ours" are statistically significant ($p \leq 0.05$).}
\end{table}

Table~\ref{tab1} presents the quantitative comparison with other methods. GNN methods such as GCN~\cite{bib18} and GraphSAGE~\cite{bib19} suffer from poor classification accuracy, especially at the subsegmental level, due to the limitations of one-hot neighborhood aggregation in capturing long-range topological relationships. HGNN methods, such as UniSAGE~\cite{bib23}, utilize hyperedges for subtree representations, while TNN~\cite{bib9} improves subtree communication via a novel sub-network architecture for enhanced subtree communication. Transformer-based methods like DeepAssign~\cite{bib20} demonstrate flexible label assignment but struggle with the complexity of subsegmental categories. AirwayFormer~\cite{bib10} employs attention mechanisms to encode tree structure priors, achieving better performance compared to other baselines. Notably, these methods do not explicitly address abnormal branches and instead treat them as a separate category, limiting their ability to handle the imbalance and undefined nature of these branches effectively.

Compared to AirwayFormer, our method achieves notable improvements. In the branch-wise evaluation, abnormal branch prediction achieves Precision, Recall, and F1 score increases of 10.0\%, 33.9\%, and 21.9\%, respectively. In the subtree-wise evaluation, SC improves by 3.1\% at the segmental level and 0.6\% at the subsegmental level, while TD decreases by 0.10 and 0.09, respectively. Clinically, these improvements enhance reliability in identifying and navigating intricate airway branches, reducing the risk of errors in bronchoscopic navigation and subsegmentectomy, particularly in cases with abnormal or highly variable branches.

\begin{figure}[!t]
\centering
\includegraphics[width=0.9 \textwidth]{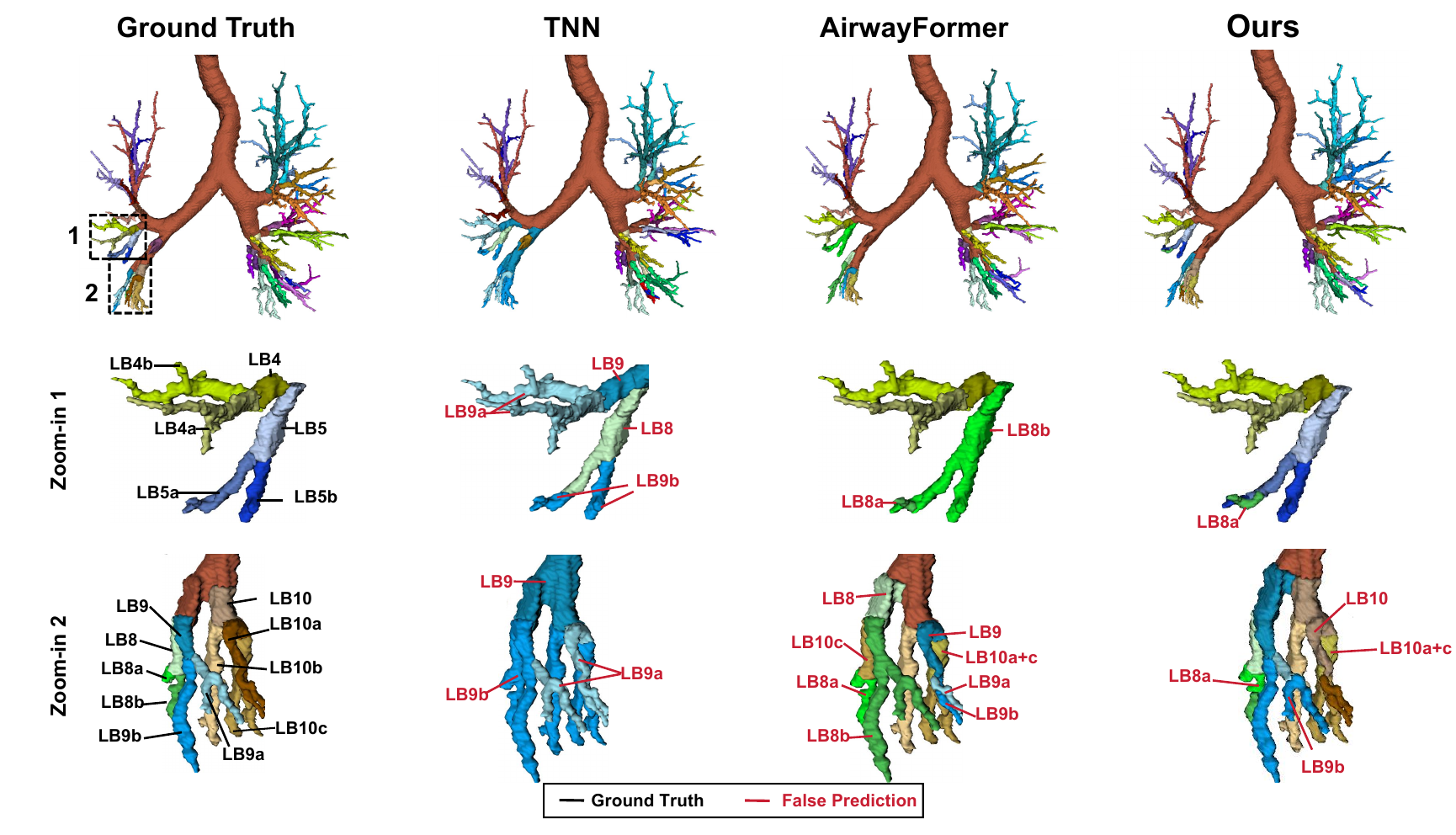}
\caption{Qualitative results at subsegmental label for deformed airways. Ground truth of branches are in black, while false predictions of branches are in red. TNN fails to generate anatomically meaningful subtrees, while AirwayFormer predicts incorrect labels. Incorporating the SSC module, our method demonstrates improved accuracy and consistency} \label{fig4}
\end{figure}

\begin{figure}[!t]
\centering
\includegraphics[width=0.9\textwidth]{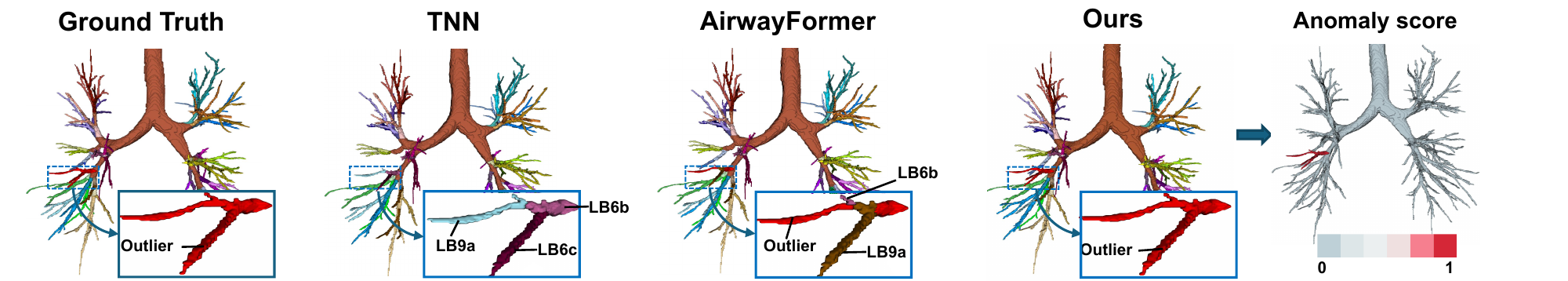}
\caption{Qualitative results at subsegmental label for abnormal branches. Competing methods misclassify abnormal branch by treating it as a separate category. Incorporating the ABS module, our method achieves superior performace} \label{fig5}
\end{figure}

Fig.~\ref{fig4} and~\ref{fig5} qualitatively demonstrate our method's superiority in different challenging scenarios. For disease-induced airway deformities(as shown in Fig.~\ref{fig4}), TNN fails to produce anatomically meaningful subtrees, misclassifying atrophied regions. AirwayFormer, focusing on node-pair relationships without subtree encoding, suffers from inconsistent predictions.  In contrast, our method leverages the SSC module, which softly integrates topological information to ensure accurate and consistent subtrees.
For abnormal branch detection (as shown in Fig.~\ref{fig5}), the absence of standardized definitions leads to prediction inconsistencies when treated equally with other categories. In contrast, our method integrates the ABS module, which enables interactions between node features and prototypes to accurately detect abnormal branches while ensuring correct classification of normal branches. 
 
\subsection{Ablation study}\label{subsec33}
\begin{table}[!t]
\centering
\setlength{\tabcolsep}{1.6pt} 
\caption{Results of ablation study of key components. PR, RC, and F1 are reported as percentages (\%). TD represents a unitless topological distance. F2C denotes network with fine-to-coarse guidance.}
\begin{tabular}{c|c|c|cccc|ccc|cc|cc}
\toprule
& \multicolumn{9}{c|}{Branch-wise} & \multicolumn{4}{c}{Subtree-wise}\\
\cmidrule(lr){2-10} \cmidrule(lr){11-14} 

 & \multicolumn{1}{c|}{Lob} & \multicolumn{1}{c|}{Seg} & \multicolumn{4}{c|}{Subsegmental} & \multicolumn{3}{c|}{Abnormal} & \multicolumn{2}{c|}{Segmental}& \multicolumn{2}{c}{Subsegmental}    \\
Method & ACC$\uparrow$ & ACC$\uparrow$ & ACC $\uparrow$& PR $\uparrow$& RC$\uparrow$ & F1 $\uparrow$& PR $\uparrow$& RC $\uparrow$& F1 $\uparrow$& SC$\uparrow$ & TD $\downarrow$& SC $\uparrow$& TD$\downarrow$ \\
\midrule
Baseline & 96.8  & 89.2 & 81.2 & 76.0 & 78.1 & 75.4 & 26.0 & 7.3 & 9.9 & 87.9 & 0.76 & 78.7 & 1.41\\
F2C & 97.2  & 90.6 & 82.7 & 78.4 & 80.0 &77.9 & 44.2 & 28.7 &31.1 & 94.3 & 0.64 &85.8 & 1.24\\
F2C + SSC & 97.4 & 90.8  & 83.2 & 78.5 & 80.2 & 78.0 &43.9 & 26.9 & 30.0 & 95.7 & 0.61& \textbf{86.7} &1.21  \\
F2C + ABS & \textbf{97.5} & 91.1 & 83.4 & 78.8 & 80.7 & \textbf{78.6} & \textbf{59.0} & 55.1 & 54.7 & 94.1 & 0.58 & 85.9 & 1.22  \\
F2C + SSC + ABS & 97.4  & \textbf{91.4}  & \textbf{83.7} & \textbf{79.1} & \textbf{80.9} & \textbf{78.6} & 55.6 & \textbf{62.9} & \textbf{54.8} & \textbf{96.8} & \textbf{0.54} & 86.3 & \textbf{1.17}\\
\bottomrule
\end{tabular}
\footnotetext{Bold values indicate the best result.}
\label{tab2}
\end{table}

Table~\ref{tab2} demonstrates the effectiveness of each component in the proposed method.
The baseline model comprises three cascaded Transformer blocks.
The fine-to-coarse (F2C) guidance encodes hierarchical dependencies across bronchial nomenclatures, and their interaction improves the classification of both normal and abnormal branches. The SSC module softly encodes subtree representation, resulting in a 1.4\% increase in segmental SC and a 0.9\% increase in subsegmental SC. The ABS module effectively enhances abnormal branch detection through the interaction between node features and prototypes, achieving 62.9\% abnormal branch recall. Additionally, it also improves the classification of normal branches by suppressing the interference of abnormal branches.
Overall, the proposed model demonstrates enhanced performance on both branch-wise and subtree-wise evaluation metrics due to the complementary contributions of its individual components.

\section{Conclusion}\label{sec4}

In this paper, an automated method for airway anatomical labeling is proposed. A soft subtree map, 
incorporating anatomically topological information, was introduced to softly encode subtree representations. 
Further, it effectively handled anomalous branches through interactions between node features and prototypes. 
The superior performance on a challenging dataset highlighting its potential for enhancing topology consistency and detecting abnormalities in clinical scenarios.

\section*{Declarations}
\begin{itemize}
\item \textbf{Funding} This work was supported in part by National Key R\&D Program of China (Grant Number: 2022ZD0212400), Natural Science Foundation of China (Grant Number: 62373243) and the Science and Technology Commission of Shanghai Municipality, China (Grant Number: 20DZ2220400), Shanghai Municipal Science and Technology Major Project (No. 2021SHZDZX0102).
\item \textbf{Competing interests} The authors have no competing interests to declare that are relevant to the content of this article.
\item \textbf{Ethics approval} This article does not contain any studies with human participants or animals performed by any of the authors.
\item \textbf{Consent for publication} The study conducted experiments based on two publicly available datasets. The access of our airway anatomical annotation can be applied for research purpose upon reasonable request.
\end{itemize}

\bibliography{bib}%

\section*{Appendix A: Sensitivity Analysis of Abnormal Score Threshold}\label{secA1}\hypertarget{secA1}{}

\begin{figure}[!h]
\centering
\includegraphics[width=0.8\textwidth]{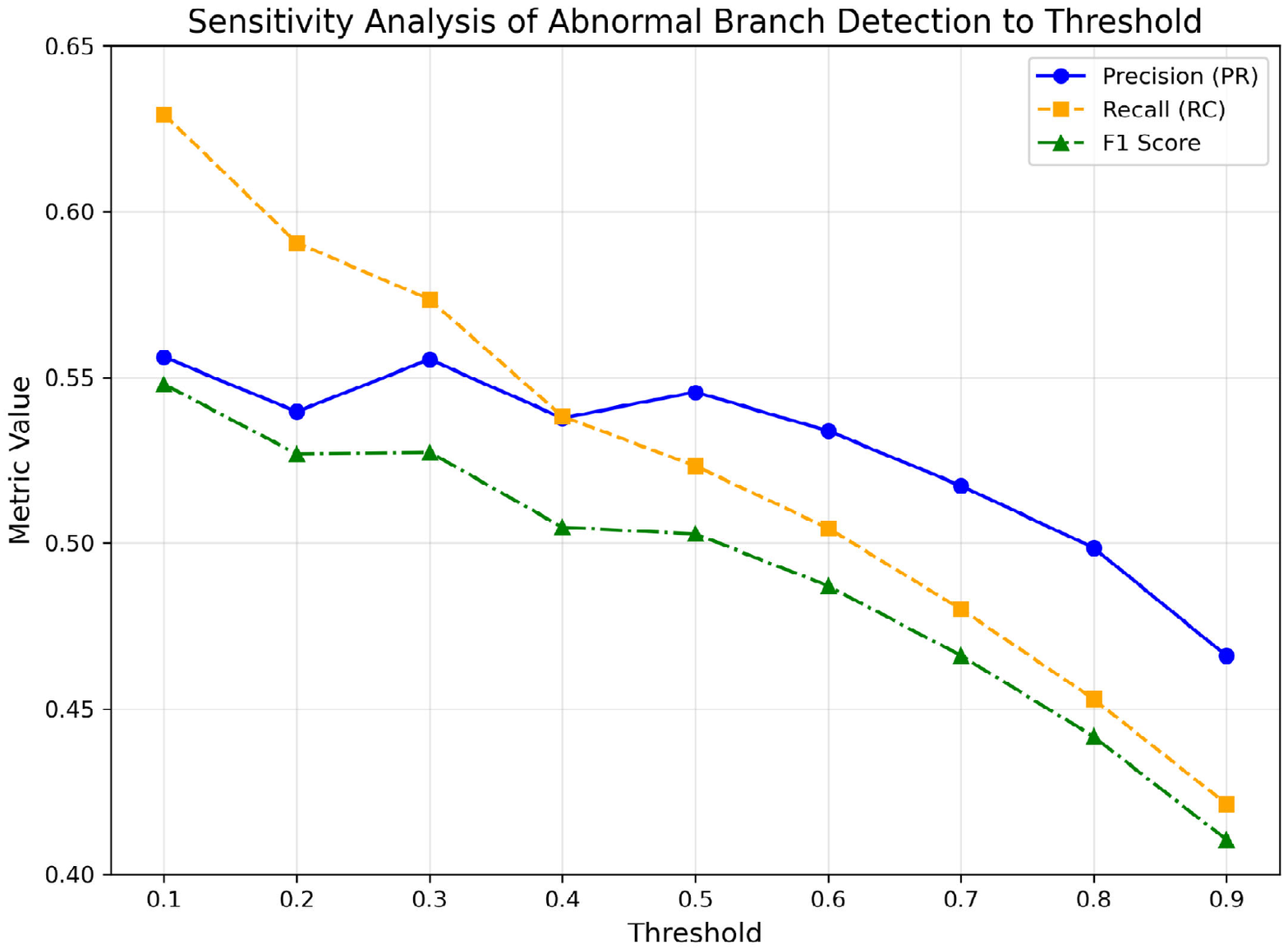}
\caption{The sensitivity analysis of abnormal branch detection performance with varying abnormal score thresholds. Metrics include Precision (PR), Recall (RC), and F1 Score. A branch is classified as abnormal if its abnormal score exceeds the threshold} \label{figA1}
\end{figure}

The appendix presents a sensitivity analysis of the abnormal score threshold in the ABS module. The analysis evaluates precision (PR), recall (RC), and F1 score across abnormal score thresholds ranging from 0.1 to 0.9, as illustrated in Fig.~\ref{figA1}. Experiments were conducted on the ATL dataset using our method.

 The result demonstrates the flexibility and reliability of ABS module under varying conditions. A threshold of 0.1 achieves the highest F1 score, balancing precision and recall. Notably, our method consistently outperforms AirwayFormer(as shown in Table~\ref{tab1}) across all metrics, particularly in Recall and F1 Score, demonstrating its superiority in abnormal branch detection.
\end{document}